\documentclass[11pt,a4paper]{article}
\usepackage[T1]{fontenc}
\usepackage[utf8]{inputenc}
\usepackage[margin=22mm]{geometry}
\usepackage{newtxtext,newtxmath}
\usepackage{microtype}
\usepackage{graphicx}
\usepackage{booktabs,tabularx,array}
\usepackage{amsmath}
\usepackage{enumitem}
\usepackage{xurl}
\usepackage{caption}
\usepackage[hidelinks]{hyperref}
\setlist{nosep,leftmargin=*}
\newcolumntype{Y}{>{\raggedright\arraybackslash}X}
\newcommand{\CT}{CNN--TCN}
\newcommand{\CL}{CNN--LIF}
\newcommand{\CA}{CNN--Transformer}
\newcommand{\ind}{\mathbf{1}}
\newcommand{\doi}[1]{\href{https://doi.org/#1}{doi: \nolinkurl{#1}}}
\title{Optical-Flow Wingbeat Counting in MuJoCo:\\A Comparison of Convolutional, Spiking,\\and Attention-Based Temporal Models}
\author{Zhang Nengbo\\[3pt]
\small School of Aerospace Engineering, Engineering Campus\\
\small Universiti Sains Malaysia, 14300 Nibong Tebal, Pulau Pinang, Malaysia\\
\small\texttt{zhangnb@student.usm.my}}
\date{}
\begin{document}
\maketitle

\begin{abstract}
Visual monitoring of flapping-wing vehicles requires distinguishing individual wingbeats from motion strength and average frequency. This paper presents a controlled MuJoCo evaluation of wingbeat counting from signed optical flow observed by virtual cameras mounted on Crazyflie vehicles. Three flapping-wing models were recorded at optical distances of 1.5 and 3.0~m, producing 1,440 clips from 240 paired scene configurations with a scene-level 3:1 training--test split. A common spatial convolutional encoder was combined with a causal temporal convolutional network, a recurrent leaky integrate-and-fire spiking network, or causal self-attention. Each model predicted phase and activity, followed by the same directed-crossing event counter. The six existing convolutional models were retained, and all twelve new models were frozen before their test predictions were generated. Exact-count accuracies at 1.5~m were 96.67\%, 95.00\%, and 96.67\%, respectively; at 3.0~m they were 94.44\%, 92.22\%, and 95.00\%. All paired scene-bootstrap intervals for differences in exact-count accuracy included zero. Seven far-distance spiking-model clips had correct totals despite event-timing mismatches, demonstrating why total-count and event-level measurements must be reported together. The results support the feasibility of causal optical-flow counting in the tested setting and identify boundary-sensitive errors. They do not establish an architecture ranking across repeated training, real-flight robustness, or hardware efficiency.
\end{abstract}

\section{Introduction}
Counting wingbeats from a camera stream provides a non-contact description of a flapping vehicle's motion. The desired output can be a cumulative number of completed repetitions, the time of each repetition, or an average frequency. These outputs impose different requirements. A frequency estimate can describe a steady oscillation while leaving individual events unresolved. A correct total can also conceal compensating missed and extra detections, or events localized at the wrong times. An observation system intended to react to individual movements therefore needs a declared event definition and an evaluation that preserves event timing.

Visual wingbeat analysis and repetition counting have established precedents. High-speed optical-flow analysis has been used to estimate the wingbeat frequency of freely flying bumblebees~\cite{santoyo2016bumblebee}. Neural repetition counting has been studied with online convolutional processing~\cite{levy2015live}, temporal self-similarity~\cite{dwibedi2020counting}, attention-based temporal correlation~\cite{hu2022transrac}, and motion-feature learning~\cite{li2024motion}. These results motivate an application-specific evaluation rather than a claim that optical flow or a particular neural architecture introduces repetition counting itself.

For micro aerial vehicles (MAVs), visually reading another vehicle's movement is also relevant to motion-mediated interaction. MoCom studied inter-MAV communication using event vision and spiking neural networks~\cite{zhang2025mocom}. Wingbeat counting is a different perceptual task: the present output is an event sequence, not a decoded navigation message. The observer--target arrangement is useful because observation distance and target geometry can be controlled while keeping the sensor input and event labels explicit. Here, three independent observer--target pairs provide three target-specific tests; there is no cross-observer information fusion or communication protocol.

This study asks three bounded questions. Can a signed-optical-flow pipeline recover wingbeat events under a reproducible simulated observation protocol? How do convolutional, recurrent spiking, and attention-based temporal modules compare when the spatial encoder structure, training budget, and event counter are held fixed? Which disagreements remain when clip totals are correct? The study contributes an inspectable evaluation workflow, a paired comparison of three temporal designs, and an analysis of event-timing and window-boundary failures. It does not introduce a new general-purpose neural architecture or claim state-of-the-art repetition-counting performance. All reported observations come from completed experiments; untested extensions are identified in the discussion.

\section{Related work}
\paragraph{Wingbeat frequency and visual repetition counting.}
Santoyo et al. combined dense optical-flow contraction and expansion with frequency analysis and state estimation to measure bumblebee wingbeats~\cite{santoyo2016bumblebee}. Their work establishes a direct precedent for using optical flow to recover periodic wing motion. Live Repetition Counting used a convolutional network and temporal accumulation for incoming video~\cite{levy2015live}. RepNet learned a temporal self-similarity representation to estimate periods across action categories~\cite{dwibedi2020counting}. TransRAC introduced multiscale temporal correlation and fine-grained repetition annotations, including videos with interruptions and inconsistent actions~\cite{hu2022transrac}. Motion-feature learning further combined RGB and motion branches to address foreground motion and background changes~\cite{li2024motion}. The present experiment uses signed horizontal and vertical flow alone as its neural input. Its comparison is internal to one controlled dataset and does not reproduce these methods' published benchmarks. In particular, published off-by-one or normalized counting errors should not be equated with the exact-count accuracy and unnormalized count MAE reported here.

\paragraph{Temporal model families.}
Causal temporal convolution is an established sequence-modeling approach~\cite{bai2018empirical}. Surrogate-gradient learning permits optimization of networks with hard spiking forward dynamics~\cite{neftci2019surrogate}. Self-attention~\cite{vaswani2017attention} and attention with linear relative-position biases~\cite{press2022train} provide another way to integrate temporal evidence. We use these components as alternative temporal modules following the same convolutional spatial encoder structure. Consequently, the spiking model is a hybrid analog-CNN/spiking model, and the attention model is a CNN/Transformer hybrid. The comparison does not assess fully spiking image encoders, pure vision Transformers, or neuromorphic hardware.

\section{Observation task and reference events}
\subsection{Simulation and image acquisition}
The observation scene was implemented in MuJoCo~3.10.0~\cite{todorov2012mujoco}. Crazyflie geometry was based on the Crazyflie~2 model in MuJoCo Menagerie~\cite{menagerie2026}. The three target assets were adapted from RLFlapping, flappy\_v2, and BIRD\_SIM~\cite{rlflapping2026,flappyv22026,birdsim2026}. Their configured nominal resting wingspans were 1.300, 0.40502, and 0.7504~m, and their baseline wing amplitudes were 0.45, 0.40, and 0.40~rad, respectively. These names identify imported geometries and joint structures; the experiments do not validate the flight performance or original aerodynamic assumptions of those projects.

Each Crazyflie carried one virtual forward-facing RGB camera and observed its corresponding target. Both body roots were nominally held at a height of 1.3~m. The camera optical center was placed either 1.5 or 3.0~m from the target root. Each image contained the assigned target at a resolution of $128\times96$ pixels, with a $45^\circ$ vertical field of view and a 60~Hz acquisition rate. Saved images were not digitally cropped or resized. The simulator ran at 2,400~Hz, giving 40 physics steps per image. Shadow rendering was disabled.

An ideal inverse-dynamics controller applied generalized forces to stabilize the bodies and track prescribed sinusoidal wing motions. For the simulated generalized position error $e$ and velocity reference $\dot q_{\mathrm{ref}}$, the inverse-dynamics acceleration request was $\ddot q_{\mathrm{ref}}+1600e+80(\dot q_{\mathrm{ref}}-\dot q)$. The resulting force was applied before each MuJoCo integration step. This establishes a controlled visual-motion experiment; it is not a demonstration that aerodynamic wing forces independently sustain flight. Observers remained nominally stationary during acquisition, and pose stabilization used simulator state separately from the visual counting pipeline.

\subsection{Paired scenes and split}
The dataset contained 240 paired scene configurations. Each configuration produced three target-camera streams at each distance, giving six clips per configuration and 1,440 clips overall. Each clip contained 180 frames with timestamps $t_i=i/60$, $i=0,\ldots,179$, after a separate 0.8~s simulation settling period. The complete collection contained 259,200 RGB frames. The scene-generation seed was 20261001.

The first designated 180 configurations were assigned to training and 60 to testing. All three targets and both distances of a configuration remained in the same split. Thus, each target--distance condition contained 180 training and 60 test clips, while each distance pooled over targets contained 540 training and 180 test clips. This was a 3:1 split by complete paired scene, not by overlapping frames or windows.

Moving-scene frequencies were drawn in the range 2.5--7.5~Hz. A scheduled subset used permutations of 3, 5, and 7~Hz; each clip retained a constant commanded frequency. Initial phase, wing amplitude, illumination, background and floor colors, and target appearance varied across scenes. Amplitudes were scaled by factors sampled from $[0.75,1.2]$ relative to the configured target amplitude. Paired distances shared these choices. Ten percent of the scenes in each split were stationary controls: 18 training and six test scenes per distance, yielding 18 static test clips when all three targets were pooled. These controls replaced the corresponding moving commands. They were retained in all primary metrics.

\subsection{Reference phase and one-event-per-cycle definition}
Reference labels came from the actual primary-wing joint position and velocity at image times. For a moving target, let $q_i$ be the signed primary-wing displacement relative to its neutral position, $\dot q_i$ its signed velocity, $A$ the configured amplitude, and $f$ the commanded frequency. The teacher phase was
\begin{equation}
\phi_i=\operatorname{unwrap}\!\left[\operatorname{atan2}\left(\frac{q_i}{A},\frac{\dot q_i}{2\pi f A}\right)\right].
\label{eq:teacher}
\end{equation}
Within the learning and evaluation pipeline, commanded frequency and amplitude were used for teacher annotation only; neither was supplied to optical-flow estimation or neural inference. They also defined the simulated motion commands. Stationary clips received a constant reference phase and zero events. Saved labels were checked against the actual joint's negative-to-positive mid-position crossings.

One wingbeat event was defined as the primary wing crossing its neutral position in the positive direction, equivalently crossing an increasing $2\pi$ phase boundary. Synchronously moving left and right wings did not count as two separate events. Event times were linearly interpolated between adjacent reference phase samples. This convention counts observed cycle landmarks, and is distinct from taking the floor of total phase advance or multiplying a commanded frequency by clip duration.

The first 30 image frames were used for history preparation. All architectures were evaluated on the same interval
\begin{equation}
\mathcal I=(t_{30},t_{179}]=(0.5,2.983\overline{3}]\ \mathrm{s},\qquad |\mathcal I|=149/60\ \mathrm{s}.
\label{eq:window}
\end{equation}
An event exactly at the interval start was excluded. No frame after the clip end was available to confirm a final candidate event.

\section{Optical-flow phase prediction and event counting}
\subsection{Inputs and shared spatial encoder}
Successive RGB images were converted to grayscale and processed by Farneb\"ack dense optical flow~\cite{farneback2003twoframe}. The fixed OpenCV parameters were pyramid scale~0.5, three levels, window size~15, three iterations, polynomial neighborhood~5, polynomial standard deviation~1.2, and flags~0. The resulting input $F_i\in\mathbb R^{96\times128\times2}$ contained signed horizontal and vertical displacements in pixels per frame. The first flow field in each clip was set to zero. Neither simulator segmentation nor label-dependent target crops were supplied to the network.

\begin{figure}[tbp]
\centering
\includegraphics[width=\linewidth]{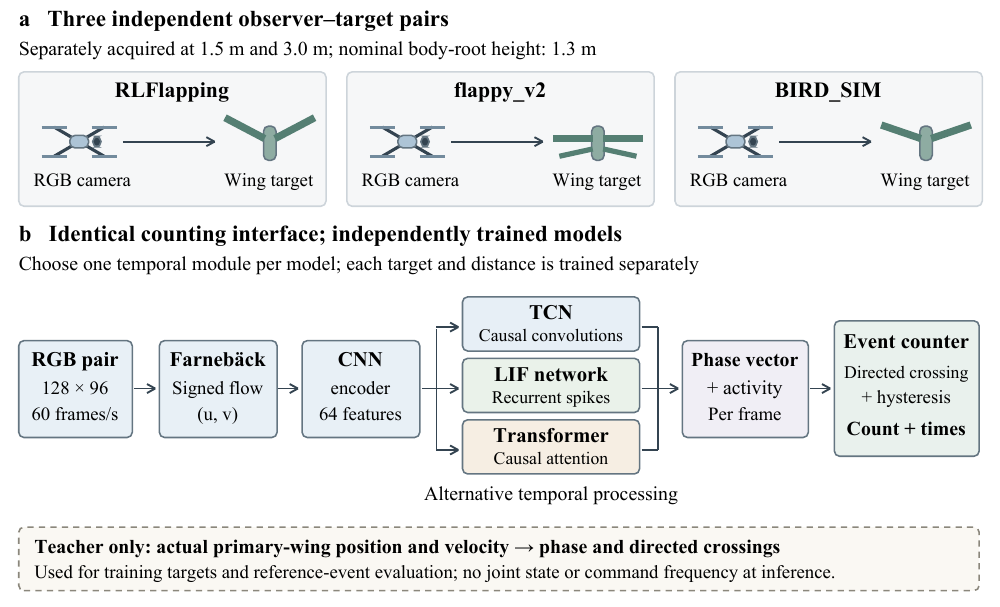}
\caption{Observation and counting information flow. Three independent camera--target pairs are evaluated at two separately acquired distances. Each model reads signed optical flow from its own RGB stream. The three temporal modules are alternatives, not an ensemble or a multi-observer fusion mechanism. They use the same spatial encoder structure with independently trained weights. Simulator phase is reserved for training and reference-event evaluation; it does not enter the inference path. The counter records an interpolated crossing time only after causal confirmation. The drawing is schematic and not to scale.}
\label{fig:pipeline}
\end{figure}

The neural pipeline is summarized in Fig.~\ref{fig:pipeline}. Flow values were transformed component-wise as
\begin{equation}
\widetilde F_i=\frac{1}{3}\operatorname{asinh}\!\left(\frac{\operatorname{clip}(F_i,-8,8)}{0.1}\right).
\end{equation}
Three $5\times5$ convolutions with stride~2 and padding~2 produced 12, 24, and 32 channels. Each stage used four-group normalization and a GELU activation. Adaptive average pooling to $3\times4$ spatial bins and a linear projection yielded a 62-dimensional vector. Two additional features, $\log(1+\operatorname{mean}\|F_i\|_2)$ and $\log(1+\max\|F_i\|_2)$, preserved raw motion-magnitude information for stationary decisions. Concatenation gave a 64-dimensional frame representation. The spatial encoder contained 51,074 trainable parameters. Its structure and initial weights were matched within a target across model families and distances; subsequent training optimized independent weights for every model.

\subsection{Alternative temporal modules}
All modules predicted an unnormalized two-dimensional phase vector $p_i=(p_i^{c},p_i^{s})$ and an activity logit $\ell_i$. The phase angle was $\hat\phi_i=\operatorname{atan2}(p_i^{s},p_i^{c})$ and the activity probability was $a_i=\sigma(\ell_i)$.

\paragraph{Convolutional temporal processing (\CT).}
Four residual temporal blocks used kernel size~3, 64 channels, and dilations 1, 2, 4, and 8. Left padding made every convolution causal. Each block applied GELU, dropout~0.05, a pointwise channel mixer, and a residual connection. A final pointwise convolution produced the three outputs. The receptive field was $1+2(1+2+4+8)=31$ frames.

\paragraph{Recurrent spiking temporal processing (\CL).}
The frame representation passed through layer normalization and a linear projection to two recurrent leaky integrate-and-fire (LIF) layers, each containing 136 neurons. For layer $l$, the per-frame update was
\begin{align}
u^{l,-}_i &= \beta u^{l,+}_{i-1}+W_l x_i^l+R_l s^l_{i-1}+b_l,\\
s_i^l &= \ind[u^{l,-}_i\geq\vartheta],\qquad
u^{l,+}_i=u^{l,-}_i-\vartheta\operatorname{sg}(s_i^l),
\end{align}
where $x_i^2=s_i^1$, $\beta=0.8$, $\vartheta=1$, and $\operatorname{sg}$ denotes a stopped reset gradient. The first-layer input was the normalized 64-dimensional encoder representation. Training used the surrogate derivative $(1+5|u^{l,-}_i-\vartheta|)^{-2}$~\cite{neftci2019surrogate}. The readout operated only on a second-layer spike trace $z_i=0.5z_{i-1}+0.5s_i^2$; no direct analog input or membrane bypass fed the output layer. Each image frame corresponded to one spiking step, with no Poisson encoding or hidden microsteps. States were zeroed for independent windows and clips. This is a dense-GPU implementation of a hybrid network, without a measured energy advantage.

\paragraph{Causal attention (\CA).}
Two pre-normalized attention blocks used width~64, four heads, a feed-forward width of~128, GELU, and dropout~0.05. Future attention positions were masked. ALiBi relative-position penalties were added to attention scores~\cite{press2022train}, with head slopes $0.25$, $0.0625$, $0.015625$, and $0.00390625$. No absolute timestamp or externally supplied phase was embedded in the features. A final layer normalization and linear readout produced phase and activity. Attention could use all preceding frames in the supplied window or clip.

\begin{table}[tbp]
\centering\small
\caption{Compared models. Parameter totals include the common spatial encoder. Approximately matched parameter totals and equal optimizer steps do not imply matched temporal context, compute, or energy.}
\label{tab:models}
\begin{tabularx}{\linewidth}{l r Y}
\toprule Model & Parameters & Temporal context \\
\midrule
\CT & 117,317 & Fixed 31-frame causal receptive field.\\
\CL & 116,077 & Recurrent state over preceding frames since reset.\\
\CA & 118,341 & Causal attention over the supplied preceding frames.\\
\bottomrule
\end{tabularx}
\end{table}

\subsection{Learning objective}
Let $m\in\{0,1\}$ indicate whether a training clip contained commanded motion and $y_i=m(\cos\phi_i,\sin\phi_i)$. The objective, evaluated after the first 30 frames of a sampled window, was
\begin{equation}
\mathcal L=\operatorname{MSE}(p,y)
+0.25\operatorname{MSE}(\Delta p,\Delta y)
+0.15\operatorname{BCEWithLogits}(\ell,m)
+0.02\operatorname{MSE}(\|p\|_2,m).
\label{eq:loss}
\end{equation}
The finite differences retained direction of phase progression. The stationary phase-vector target was zero, so its arbitrary constant teacher phase did not contribute a motion target. The loss components were kept identical across model families. Their individual benefit was not established by an ablation experiment in this study.

\subsection{Causal event counter}
The same stateful counter processed every model's outputs. A sample was active when $a_i>0.5$. An active phase observation required $\|p_i\|_2\geq0.25$ and finite values. Finite inactive predictions were available stationary decisions even when the phase vector was zero. Unavailable observations, inactivity, or timestamp gaps exceeding $1.5/60$~s cleared phase tracking while preserving the accumulated count.

Within an active segment, wrapped angular differences were integrated to track increasing phase. A candidate boundary $2\pi k$ became armed after the estimated phase reached its negative hysteresis side, $2\pi k-0.1$. A positive zero crossing created a candidate with an interpolated crossing time. It was confirmed only when phase reached $2\pi k+0.1$. Confirmed events had to be separated by at least 0.08~s, and the same boundary could not be emitted twice. The cumulative count increased at confirmation time, whereas the saved crossing timestamp was used for event matching. The latter is retrospective interpolation of an already confirmed event, not permission to display a count before it became available.

The counter never filled missing observations by extrapolating commanded frequency. Likewise, a candidate near the end of the clip that had not reached positive hysteresis was not counted. At the evaluation start, the counter initialized from the current phase rather than importing a previously armed event from the warm-up period. This deliberate boundary convention can lose a reference event very close to either endpoint and is retained in the reported errors.

\section{Training and evaluation protocol}
\subsection{Training budget and retained baseline}
Each target and distance had independently trained models, giving 18 models across the three temporal families. Every model used 60 epochs, 24 optimizer steps per epoch, batch size~6, and 96-frame windows. An epoch comprised 24 minibatches sampled with replacement, rather than a full traversal of the training partition; the total was 1,440 updates and 8,640 sampled windows per model. AdamW used an initial learning rate of $10^{-3}$, weight decay $10^{-4}$, cosine decay to $5\times10^{-5}$, and gradient-norm clipping at~5. Each sampled window received one positive flow-magnitude gain drawn uniformly from $[0.9,1.1]$. The full 180-clip training partition was eligible for sampling. There was no validation-based model selection, early stopping, or architecture-specific hyperparameter search; epoch~60 was retained.

Seeds were 2026091603, 2026091703, and 2026091803 for RLFlapping, flappy\_v2, and BIRD\_SIM, respectively, with the same target seed used across distances and architectures. Initialization hashes verified identical starting spatial encoders. Recorded sampling hashes verified identical clip/window selection schedules. Gain distributions were the same, but gain values were not guaranteed to be identical because the architectures consumed the shared CUDA random-number stream differently. TF32 was disabled and deterministic settings were requested with warnings rather than hard failures. Adaptive-pooling and attention backward operations can retain numerical nondeterminism; bitwise retraining reproducibility is not claimed.

The six \CT\ models and their evaluation reports had been completed in the preceding optical-flow experiment. They were audited and retained byte-for-byte, rather than retrained or reselected. Twelve \CL\ and \CA\ models were then trained under a written comparison protocol. All twelve checkpoints were frozen, and hashes of all 18 checkpoints were checked, before any new test prediction was generated. No test-based threshold adjustment or checkpoint selection followed these predictions. Training used Python~3.10.20, PyTorch~2.11.0+cu128, OpenCV~4.13.0, and two NVIDIA GeForce RTX~5060~Ti GPUs.

The test scenes had previously been used during development of the optical-flow approach. Consequently, this is a reused fixed-split development comparison, not evaluation on a newly collected external holdout. Each model has one training realization. Near- and far-distance models were trained separately, so the comparison does not demonstrate transfer of one trained model between distances.

\subsection{Count and event metrics}
For $M$ clips, let $N_j$ and $\hat N_j$ denote the reference and predicted event counts within Eq.~\eqref{eq:window}. Exact-count accuracy and unnormalized count mean absolute error were
\begin{equation}
\mathrm{Acc}_{\mathrm{exact}}=\frac{1}{M}\sum_j\ind[\hat N_j=N_j],\qquad
\mathrm{MAE}_{\mathrm{count}}=\frac{1}{M}\sum_j|\hat N_j-N_j|.
\end{equation}
All clips remained in the denominator, including stationary clips and clips with unavailable observations. The unit of count MAE is events per clip.

Predicted and reference event timestamps were matched one-to-one within a tolerance of $2/60$~s (two camera frames). Matching first maximized the number of pairs and then minimized total timing error among tied solutions. Matched events were true positives (TP); unmatched predictions were false positives (FP); unmatched references were false negatives (FN). Precision, recall, and F1 were computed from pooled event totals, with $\mathrm{F1}=2\mathrm{TP}/(2\mathrm{TP}+\mathrm{FP}+\mathrm{FN})$. We did not average per-clip F1, which would allow zero-event static clips to inflate an event metric. FP and FN describe unmatched events under this tolerance and can arise from timing displacement even when clip totals agree.

\subsection{Paired uncertainty and implementation checks}
The 180 test clips per distance came from 60 scene configurations with three targets each. We therefore resampled paired scene clusters rather than treating all 180 clips as independent. Ten thousand bootstrap replicates used seed~2026091609, retaining all three targets and all architecture/distance observations of a selected scene together. The same resampled scene indices were used for model contrasts. Equal-tailed 95\% intervals were the 2.5th and 97.5th percentiles. These are individual, unadjusted intervals conditional on the fixed trained weights; they do not quantify variation across retraining or support simultaneous claims across all contrasts.

Separate synthetic checks verified that future inputs did not change earlier outputs, that the LIF forward path emitted binary spikes with reset, and that chunked LIF inference with carried state matched full-sequence inference. They also checked encoder initialization and frozen-source integrity. Each model received an additional 180-frame all-zero optical-flow control. These checks establish specific implementation properties, not robustness to untested camera motion or real-world backgrounds.

\section{Results}
\subsection{Exact counts and event-level performance}
At 1.5~m, \CT\ and \CA\ each counted 174 of 180 clips exactly, while \CL\ counted 171 exactly (Table~\ref{tab:overall}). At 3.0~m, the corresponding counts were 170, 171, and 166 for \CT, \CA, and \CL. Thus, the far-distance advantage of \CA\ over \CT\ was one correctly counted clip, or 0.56 percentage points. Count MAE ranged from 0.0333 to 0.0778 events per clip. These figures describe the configured models within the tested conditions.

\begin{table}[tbp]
\centering\small\setlength{\tabcolsep}{4pt}
\caption{Main results, pooling three target types. Each row contains 180 clips from 60 paired scenes, including 18 static clips. Accuracy and its interval are percentages. Intervals use paired scene-cluster bootstrap resampling; F1 uses micro event totals.}
\label{tab:overall}
\begin{tabular}{l l r r r r r}
\toprule Distance & Model & Exact & Accuracy & 95\% interval & Count MAE & Event F1\\
\midrule
1.5 & CNN--TCN & 174/180 & 96.67 & [93.89, 98.89] & 0.0333 & 0.998470 \\
1.5 & CNN--LIF & 171/180 & 95.00 & [91.67, 97.78] & 0.0500 & 0.997705 \\
1.5 & CNN--Transformer & 174/180 & 96.67 & [93.89, 98.89] & 0.0333 & 0.998470 \\
3.0 & CNN--TCN & 170/180 & 94.44 & [91.11, 97.22] & 0.0556 & 0.997449 \\
3.0 & CNN--LIF & 166/180 & 92.22 & [87.78, 96.11] & 0.0778 & 0.992861 \\
3.0 & CNN--Transformer & 171/180 & 95.00 & [91.67, 97.78] & 0.0500 & 0.997706 \\
\bottomrule
\end{tabular}
\end{table}

There were 1,964 reference events per distance, identical by construction across the paired observations (Table~\ref{tab:events}). Near-distance \CT\ and \CA\ each produced 1,958 matched events with no unmatched predictions. Far-distance \CL\ produced 1,947 matched events, 11 unmatched predictions, and 17 unmatched references, yielding F1~0.992861. Its exact-count accuracy remained 92.22\%, illustrating that a high fraction of correct totals does not summarize all event-timing errors.

\begin{table}[tbp]
\centering\small\setlength{\tabcolsep}{4pt}
\caption{Reference and predicted event totals and one-to-one matching. P and R denote micro precision and recall in percent. An unmatched event may reflect a timing error, a missed cycle, or an extra detected cycle.}
\label{tab:events}
\begin{tabular}{l l r r r r r r r}
\toprule Distance & Model & Ref. & Pred. & TP & FP & FN & P & R\\
\midrule
1.5 & CNN--TCN & 1964 & 1958 & 1958 & 0 & 6 & 100.000 & 99.695 \\
1.5 & CNN--LIF & 1964 & 1957 & 1956 & 1 & 8 & 99.949 & 99.593 \\
1.5 & CNN--Transformer & 1964 & 1958 & 1958 & 0 & 6 & 100.000 & 99.695 \\
3.0 & CNN--TCN & 1964 & 1956 & 1955 & 1 & 9 & 99.949 & 99.542 \\
3.0 & CNN--LIF & 1964 & 1958 & 1947 & 11 & 17 & 99.438 & 99.134 \\
3.0 & CNN--Transformer & 1964 & 1959 & 1957 & 2 & 7 & 99.898 & 99.644 \\
\bottomrule
\end{tabular}
\end{table}

\subsection{Target-specific behavior}
Table~\ref{tab:targets} retains all target-specific results. Near-distance flappy\_v2 was counted exactly in all 60 clips by \CT\ and \CA. This finite result is specific to that model, geometry, and test partition. At the far distance, the largest event-F1 reduction occurred for \CL\ on flappy\_v2: 54 of 60 totals were exact, but event F1 was 0.98471. The other two far-distance \CL\ target conditions had F1 above 0.995. The observed difference therefore did not occur uniformly across target types. Because the assets have different physical wingspans, equal observation distances do not equalize their projected image sizes; target-specific differences cannot be attributed to temporal processing alone.

\begin{table}[tbp]
\centering\small\setlength{\tabcolsep}{4pt}
\caption{Target-specific results. Every row includes 60 clips, of which six are static. Distance is in meters, accuracy in percent, and count MAE in events per clip.}
\label{tab:targets}
\begin{tabular}{l l l r r r r}
\toprule Distance & Model & Target & Exact & Accuracy & Count MAE & Event F1\\
\midrule
1.5 & CNN--TCN & RLFlapping & 58/60 & 96.67 & 0.033 & 0.99840 \\
1.5 & CNN--TCN & flappy\_v2 & 60/60 & 100.00 & 0.000 & 1.00000 \\
1.5 & CNN--TCN & BIRD\_SIM & 56/60 & 93.33 & 0.067 & 0.99707 \\
1.5 & CNN--LIF & RLFlapping & 56/60 & 93.33 & 0.067 & 0.99680 \\
1.5 & CNN--LIF & flappy\_v2 & 58/60 & 96.67 & 0.033 & 0.99847 \\
1.5 & CNN--LIF & BIRD\_SIM & 57/60 & 95.00 & 0.050 & 0.99780 \\
1.5 & CNN--Transformer & RLFlapping & 57/60 & 95.00 & 0.050 & 0.99760 \\
1.5 & CNN--Transformer & flappy\_v2 & 60/60 & 100.00 & 0.000 & 1.00000 \\
1.5 & CNN--Transformer & BIRD\_SIM & 57/60 & 95.00 & 0.050 & 0.99780 \\
3.0 & CNN--TCN & RLFlapping & 57/60 & 95.00 & 0.050 & 0.99760 \\
3.0 & CNN--TCN & flappy\_v2 & 58/60 & 96.67 & 0.033 & 0.99847 \\
3.0 & CNN--TCN & BIRD\_SIM & 55/60 & 91.67 & 0.083 & 0.99633 \\
3.0 & CNN--LIF & RLFlapping & 57/60 & 95.00 & 0.050 & 0.99760 \\
3.0 & CNN--LIF & flappy\_v2 & 54/60 & 90.00 & 0.100 & 0.98471 \\
3.0 & CNN--LIF & BIRD\_SIM & 55/60 & 91.67 & 0.083 & 0.99633 \\
3.0 & CNN--Transformer & RLFlapping & 57/60 & 95.00 & 0.050 & 0.99760 \\
3.0 & CNN--Transformer & flappy\_v2 & 57/60 & 95.00 & 0.050 & 0.99770 \\
3.0 & CNN--Transformer & BIRD\_SIM & 57/60 & 95.00 & 0.050 & 0.99780 \\
\bottomrule
\end{tabular}
\end{table}

\subsection{Paired differences}
All 95\% intervals for pairwise differences in exact-count accuracy included zero (Table~\ref{tab:paired}). For example, the far-distance \CA\ minus \CT\ difference was $+0.56$ percentage points with interval $[-1.67,+2.78]$. The point estimates therefore do not establish a stable exact-count ranking, and an interval spanning zero is not an equivalence test.

The far-distance event-F1 differences between \CL\ and the other two models had individual intervals excluding zero: \CL\ minus \CT\ was $-0.004588$ with interval $[-0.008936,-0.001059]$, and \CA\ minus \CL\ was $+0.004845$ with interval $[+0.001285,+0.009106]$. These are conditional, unadjusted contrasts over the same test scenes. They should not be generalized to all training seeds or used as a claim that spiking temporal processing is intrinsically inferior.

\begin{table}[tbp]
\centering\footnotesize\setlength{\tabcolsep}{3pt}
\caption{Paired model differences, written as first minus second, with individual 95\% scene-bootstrap intervals. Accuracy differences are percentage points; F1 differences use the 0--1 scale. The intervals do not include training-seed variability and are not corrected for multiple comparisons.}
\label{tab:paired}
\begin{tabular}{l l r r}
\toprule Distance & Contrast & $\Delta$ exact accuracy [95\% interval] & $\Delta$ event F1 [95\% interval]\\
\midrule
1.5 & CNN--LIF $-$ CNN--TCN & -1.67 [-4.44, +1.11] & -0.000766 [-0.002098, +0.000521] \\
1.5 & CNN--Transformer $-$ CNN--TCN & +0.00 [-1.67, +1.67] & +0.000000 [-0.000755, +0.000753] \\
1.5 & CNN--Transformer $-$ CNN--LIF & +1.67 [-0.56, +4.44] & +0.000766 [-0.000264, +0.001938] \\
3.0 & CNN--LIF $-$ CNN--TCN & -2.22 [-6.67, +1.67] & -0.004588 [-0.008936, -0.001059] \\
3.0 & CNN--Transformer $-$ CNN--TCN & +0.56 [-1.67, +2.78] & +0.000257 [-0.000810, +0.001371] \\
3.0 & CNN--Transformer $-$ CNN--LIF & +2.78 [-1.11, +7.22] & +0.004845 [+0.001285, +0.009106] \\
\bottomrule
\end{tabular}
\end{table}

\subsection{Stationary controls and observation availability}
No architecture produced a false event in the 18 stationary test clips at either distance. All 18 additional zero-flow controls also produced zero counts (Table~\ref{tab:controls}). Three architecture--clip records had at least one unavailable observation: one near \CT\ record, one far \CL\ record, and one far \CA\ record. They were retained in every primary metric. Because static clips are easier to count as zero, Table~\ref{tab:controls} also reports accuracy on the 162 moving clips per distance.

\begin{table}[tbp]
\centering\small\setlength{\tabcolsep}{4pt}
\caption{Controls and availability. Static errors are incorrectly counted static clips out of 18; zero-flow failures are models producing a nonzero count out of three. Unavailable records were not excluded. Moving accuracy uses only 162 moving clips and is expressed in percent.}
\label{tab:controls}
\begin{tabular}{l l r r r r}
\toprule Distance & Model & Static errors & Zero-flow failures & Unavailable & Moving accuracy\\
\midrule
1.5 & CNN--TCN & 0/18 & 0/3 & 1 & 96.30 \\
1.5 & CNN--LIF & 0/18 & 0/3 & 0 & 94.44 \\
1.5 & CNN--Transformer & 0/18 & 0/3 & 0 & 96.30 \\
3.0 & CNN--TCN & 0/18 & 0/3 & 0 & 93.83 \\
3.0 & CNN--LIF & 0/18 & 0/3 & 1 & 91.36 \\
3.0 & CNN--Transformer & 0/18 & 0/3 & 1 & 94.44 \\
\bottomrule
\end{tabular}
\end{table}

\subsection{Post-hoc boundary and timing analysis}
Saved event lists were inspected without changing checkpoints, predictions, or counting thresholds. Every unmatched event for \CT, \CA, and near-distance \CL\ lay within one camera frame of an evaluation endpoint (Table~\ref{tab:diagnostics}). This concentration is consistent with the counter's requirement to arm after the interval begins and to confirm before the clip ends. The preceding \CT\ inspection distinguished initial unarmed crossings, final unconfirmed candidates, and final predicted phases that had not yet crossed. These are part of the current protocol's measured error, rather than observations removed from evaluation.

Far-distance \CL\ additionally had seven unmatched reference events and seven unmatched predicted events outside the endpoint neighborhoods. Seven far-distance \CL\ clips had equal predicted and reference totals despite nonzero FP and FN. Correct totals in those cases did not imply accurate event localization. The stored timing mismatches, rather than total-count accuracy alone, identified this failure mode. This post-hoc description does not isolate its architectural cause and was not used for test-driven revisions.

\begin{table}[tbp]
\centering\small\setlength{\tabcolsep}{4pt}
\caption{Post-hoc error locations. Boundary FN and FP are counts within $1/60$~s of either endpoint divided by all FN or FP in that condition. A zero denominator means there were no errors of that type. The last column counts clips whose totals are exact but whose event matching has an error.}
\label{tab:diagnostics}
\begin{tabular}{l l r r r}
\toprule Distance & Model & Boundary FN / all FN & Boundary FP / all FP & Exact but mismatched\\
\midrule
1.5 & CNN--TCN & 6/6 & 0/0 & 0 \\
1.5 & CNN--LIF & 8/8 & 1/1 & 0 \\
1.5 & CNN--Transformer & 6/6 & 0/0 & 0 \\
3.0 & CNN--TCN & 9/9 & 1/1 & 0 \\
3.0 & CNN--LIF & 10/17 & 4/11 & 7 \\
3.0 & CNN--Transformer & 7/7 & 2/2 & 0 \\
\bottomrule
\end{tabular}
\end{table}

\section{Discussion and limitations}
The main empirical finding is that three modest temporal models can support causal wingbeat counting from signed optical flow in this controlled task, while exact totals and correctly timed events remain distinct outcomes. A second finding is the strong influence of evaluation endpoints on most observed errors. With a count interval of only 149/60~s, losing one event near a boundary makes an otherwise accurate clip fail exact-count evaluation. Changing the boundary protocol could change the reported percentage. A future protocol should specify whether state may carry across windows, whether delayed confirmation after a reporting boundary is allowed, and how confirmation delay is scored, before evaluating new models.

The comparison controls several factors but does not isolate an architecture effect. Spatial encoder structure, initial encoder weights, scene split, sampled windows, loss, optimizer schedule, and counter thresholds were matched. Parameter totals were close. However, temporal context differed, random augmentation draws were not identical, and each architecture used one fixed hyperparameter configuration and one training realization per condition. The CNN baseline was reused after prior evaluation. These choices support a practical development comparison, not a universal ranking or a test of optimally tuned representatives of each model family.

The simulation also limits the application claim. Frequencies were constant within clips, cameras were nominally stationary, and visual variation consisted mainly of controlled appearance and illumination changes. There were no systematic tests of observer ego motion, continuous frequency change, long-duration drift, occlusion, motion blur, dropped frames, or realistic camera noise. The near and far models were trained separately; no target type, distance, or background family was withheld as an unseen domain. Ideal generalized-force stabilization does not establish aerodynamic free-flight control. Three separate observer streams likewise do not establish cooperative perception or communication.

Causal information flow is necessary for an online implementation, but it is not a measurement of real-time deployment. The current evaluation processed saved clips, and no end-to-end camera-to-event latency, sustained onboard throughput, or hardware energy was measured. The spiking path was implemented with dense GPU operations after an analog CNN, so an energy-efficiency claim cannot be inferred from its binary spikes. A deployment study would have to include optical-flow computation, spatial encoding, temporal state handling, and event confirmation latency in the same measurement.

Finally, this work contains an internal architecture comparison, not a reproduced comparison against established repetition-counting methods or a classical peak/phase-tracking baseline. The relevant literature supplies prior approaches and context, not directly comparable scores on this dataset. A stronger follow-up would lock a new scene-disjoint holdout, repeat training across seeds, compare appropriate causal baselines, and test real video of a flapping mechanism before making broader robotic claims. The present version preserves its completed measurements and explicit failure cases as a starting point for those tests.

\section{Conclusion}
A controlled MuJoCo workflow was developed to count individual wingbeat events from signed optical flow using alternative convolutional, recurrent spiking, and attention-based temporal modules. Near-distance exact-count accuracy ranged from 95.00\% to 96.67\%, and far-distance accuracy from 92.22\% to 95.00\%, under the same fixed event counter. Paired intervals did not establish an exact-count architecture ranking. Event-level analysis revealed both endpoint-sensitive errors and seven far-distance spiking-model clips with correct totals but inaccurate event matching. These results support an inspectable simulation study of wingbeat counting and motivate evaluating event timing, confirmation, and cumulative counts together when extending the task to moving cameras and real platforms.

\section*{Data, code, and model availability}
The accompanying source submission includes the experiment protocol, aggregate statistics, and per-clip count and event-matching records as ancillary files. The statistical records identify the paired scenes so that aggregate count metrics and paired comparisons can be checked. The complete RGB/optical-flow arrays and trained checkpoints were retained locally for this study and are not deposited in a public repository in this version. Upstream model sources are cited separately; their meshes are not redistributed with the paper. No public release of assets without established redistribution permissions is implied. The availability scope limits independent regeneration of the full image-to-prediction pipeline from the paper package alone.

\section*{AI assistance}
AI assistance was used for implementation support, analysis-script development, literature organization, and English drafting. The numerical tables were generated from saved per-clip evaluation records and checked against frozen experiment artifacts. The schematic is an original vector drawing of the implemented information flow.

\begingroup
\small
\setlength{\parskip}{0pt}
\bibliographystyle{unsrturl}
\bibliography{references}
\endgroup
\end{document}